\documentclass[conference]{IEEEtran}
\usepackage{graphics} % for pdf, bitmapped graphics files
\usepackage{epsfig} % for postscript graphics files
\usepackage{graphicx}
\usepackage{cuted}
\usepackage{flushend}
\usepackage{multirow}
\usepackage{mathptmx}
\usepackage{pbox}
\usepackage{csquotes}
\usepackage{array}
\usepackage{booktabs}
\usepackage{threeparttable}
\usepackage{wrapfig}
\usepackage{float}
\usepackage{caption}
\usepackage{graphicx}
\usepackage{array}
\usepackage{svg}
\usepackage{cite}
\usepackage{amsmath,amssymb,amsfonts}
\usepackage{algorithmic}
\usepackage{textcomp}
\usepackage{xcolor}
\def\BibTeX{{\rm B\kern-.05em{\sc i\kern-.025em b}\kern-.08em
    T\kern-.1667em\lower.7ex\hbox{E}\kern-.125emX}}

\usepackage{fancyhdr}
\begin{document}

\title{Stacked Ensemble of Fine-Tuned CNNs for Knee Osteoarthritis Severity Grading
}

\author{\textsuperscript{1}A. Gupta, \textsuperscript{1}J. Kaur, \textsuperscript{1}T. Doshi,  \textsuperscript{1}T. Sharma, \textsuperscript{2}N. K. Verma, and \textsuperscript{3}S. Vasikarla
\\
\textsuperscript{1}\textit{Indian Institute of Technology Guwahati, India}\\
\textsuperscript{2}\textit{Indian Institute of Technology Kanpur, India}\\
\textsuperscript{3}\textit{California State University, Northridge, CA, USA}\\
Email: adarsh.gupta@iitg.ac.in, japleen@iitg.ac.in, t.doshi@iitg.ac.in,\\ teena@iitg.ac.in, nishchal@iitk.ac.in, shantaramv@hotmail.com
}

\maketitle

\begin{abstract}
Knee Osteoarthritis (KOA) is a musculoskeletal condition that can cause significant limitations and impairments in daily activities, especially among older individuals. To evaluate the severity of KOA, typically, X-ray images of the affected knee are analyzed, and a grade is assigned based on the Kellgren-Lawrence (KL) grading system, which classifies KOA severity into five levels, ranging from 0 to 4. This approach requires a high level of expertise and time and is susceptible to subjective interpretation, thereby introducing potential diagnostic inaccuracies. To address this problem a stacked ensemble model of fine-tuned Convolutional Neural Networks (CNNs) was developed for two classification tasks: a binary classifier for detecting the presence of KOA, and a multiclass classifier for precise grading across the KL spectrum. The proposed stacked ensemble model consists of a diverse set of pre-trained architectures, including MobileNetV2, You Only Look Once (YOLOv8), and DenseNet201 as base learners and Categorical Boosting (CatBoost) as the meta-learner. This proposed model had a balanced test accuracy of 73\% in multiclass classification and 87.5\% in binary classification, which is higher than previous works in extant literature.
\end{abstract}


\section{Introduction}
\label{sec:Introduction}
Knee Osteoarthritis (KOA) \cite{KOA} is a degenerative musculoskeletal joint disease in which the knee cartilage breaks down over time. It is characterized by joint pain, swelling, and, in extreme cases, mobility limitations. Approximately 13\%, 10\%, and 40\% of women, men aged above 60 years of age, and people above 70 years of age, respectively, have symptomatic KOA \cite{KOAStats}. X-ray scans are used for diagnosing KOA due to their safety and cost-effectiveness \cite{XRay}. 

The Kellgren and Lawrence (KL) grading system categorizes the progress of KOA into 5 grades  \cite{KL}. Grade 0 signifies \enquote*{No Disease}, Grade 1 signifies \enquote*{Doubtful}, Grade 2 signifies \enquote*{Minimal} revealing the presence of mild KOA with the existence of possible Joint Space Narrowing (JSN), Grade 3 signifies \enquote*{Moderate} KOA with noticeable JSN and lastly, Grade 4 denotes \enquote*{Severe} KOA. 

The only available treatments for KOA are behavioral interventions, which, at best, may only slow down the course of the disease  \cite{treatment}. Thus, the best way to tackle KOA and prevent future disability is early detection. Due to these reasons and the rising prevalence of KOA, there is an urgent need for automated and efficient approaches for KOA detection and classification. This makes using Artifical Intelligence-based (AI-based) diagnosis systems extremely crucial to aid medical professionals in making sound decisions, as AI-based systems can help achieve objective and reproducible results quickly.

\begin{figure*}
    \centering
    \includegraphics[width=0.99\linewidth]{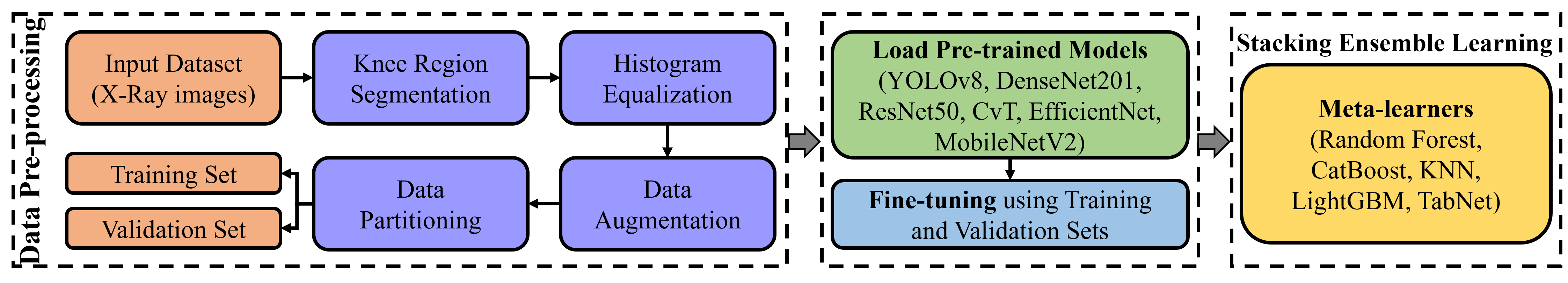}
    \caption{Workflow of the proposed methodology: Data Pre-processing, fine-tuning, and meta-learning.}
    \label{fig:workflow of the proposed methodology}
\end{figure*}

\subsection{Related Works}

In recent years, deep learning-based image classification models for disease diagnosis have gained popularity for various diseases, including Alzheimer's, breast cancer, pneumonia, and heart failure \cite{disease}.

In 2016, Antony \textit{et al.} \cite{antony}, were the first to use a fully Convolutional Neural Network (CNN) to assess KOA severity, which had an accuracy of 53.4\%. In 2018, Tiulpin \textit{et al.} \cite{tiulpin} introduced a technique based on deep Siamese CNNs, which employs Contrastive Loss  \cite{siamese} to gauge the similarity between pairs of images within a dataset, achieving a test accuracy of 66.71\%.

In 2019, Chen \textit{et al.} \cite{chen} developed a model based on two CNNs, and the best-performing one was chosen for classification. In his work, a specialized You Only Look Once (YOLOv2) \cite{YOLO2} network was employed to classify the X-ray images. Tiulpin \textit{et al.} \cite{Tiulpin} achieved 0.98 value for Area under the Receiver Operating Characteristic Curve (AUROC/AUC), using a CNN ensemble network of 50 layers. In 2022, Yadav \textit{et al.} \cite{yadav} proposed a highly effective Scaling-down Fusion Network (SFNet) model, and Lau \textit{et al.} \cite{lau} suggested using an Xception \cite{xcep} network trained on the ImageNet dataset \cite{dataset} and a Knee X-ray image dataset. All works on using Machine Learning (ML) and Deep Learning (DL) techniques together showed excellent results for binary classification (detection of KOA). However, they did not prove to be very accurate for multiclass classification (grading of KOA). 

Mohammed \textit{et al.} \cite{moham} trained six CNN models on the Osteoarthritis Initiative (OAI)\cite{OAI} Dataset. %They experimented with the effect of varying the number of classes on the classification accuracy by creating 3 datasets from the original OAI dataset. 
They achieved the highest test accuracy of 69\% in multiclass classification, 83\% in binary classification, and 89\% in 3-class classification.

This paper aims to build on previous works in an attempt to increase the classification accuracy by using a stacked ensemble \cite{stack} based learning technique using a class-weighted loss function on the following state-of-the-art (SOTA) CNNs - MobileNetV2  \cite{MobileNetv2}, ResNet50  \cite{resnet}, EfficientNet  \cite{efficientnet}, YOLOv8  \cite{YOLO}, DenseNet201  \cite{densenet}, and Convolutional Vision Transformer (CvT) \cite{cvt}, for both the detection and classification of KOA.

\subsection{Contributions}

Herein, the problem of automated KOA detection and grading has been solved by proposing a novel stacked CNN ensemble model with MobileNetV2, YOLOv8, and DenseNet201 as base learners and Categorical Boosting (CatBoost) \cite{catboost} as the meta-learner. The contributions can be encapsulated as below:
\begin{enumerate}
    \item This paper uses a class-weighted loss function to address the significant issue of class imbalance in the dataset. The objective is to assign a higher penalty for the misclassification of the minority class to prevent the model from becoming biased.
    \item This paper uses various fine-tuned SOTA CNNs, including YOLOv8, MobileNetV2, and DensNet201, as base learners to achieve higher classification accuracy when using the meta-learners.
    \item This paper experimented with various meta-learners such as CatBoost, RandomForest \cite{randomforest}, K-Nearest Neighbours (KNN) \cite{knn}, Light Gradient-Boosting Machine (LightGBM) \cite{lgbm} and, TabNet \cite{tabnet}. After hyperparameter tuning all meta-learners, CatBoost was chosen as the best meta-learner.
\end{enumerate}

Furthermore, the paper includes Section \ref{sec: Proposed Methodology} to present the suggested approach, Section \ref{sec: Experimental Results, Validations, and Discussions} details the empirical findings, validations, and discussions, and at the end, Section \ref{sec:Conclusion and Future Scopes} covers the concluding mentions along with the purview of future possibilities.

\section{Proposed Methodology}
\label{sec: Proposed Methodology}

The proposed methodology for creating the final model has three parts: Data Pre-processing, Fine-tuning the pre-trained CNNs using a class-weighted loss function, and training a meta-learner to complete the stacked ensemble. Fig. \ref{fig:workflow of the proposed methodology} illustrates the pictorial workflow of the introduced methodology.

\subsection{Data Pre-processing}

% \begin{figure} % Use figure instead of wrapfigure to avoid text wrapping
%     \centering
%     \includegraphics[width=0.49\textwidth]{preformatting.png} 
%     \caption{Data Pre-processing pipeline example}
%     \label{fig:example_pre}
% \end{figure}

The Knee X-ray images of the OAI dataset were segmented to isolate the region of the Knee that contains the most critical features required for KOA detection and grading.

After segmentation, the sharpness and contrast of the images needed to be improved for faster fine-tuning of the CNNs. This was done by applying Contrast Limited Adaptive Histogram Equalization (CLAHE) \cite{CLAHE}. CLAHE is an improved histogram equalization strategy over Adaptive Histogram Equalization (AHE). AHE is a common method to increase the sharpness, contrast, and edge definition in different parts of the image. It computes histograms in distinct sections of the image and uses this information to redistribute the luminance values of the image \cite{clahe enhancement}. CLAHE operates on smaller regions of the images and then combines neighboring regions by binary interpolation, which can help reduce noise amplification that AHE can lead to \cite{clahe noise}.

Following the application of segmentation and CLAHE, the data is augmented to enhance the \enquote{size} and \enquote{diversity} of the dataset and subsequently resized to meet the input dimensions required by the specific model during fine-tuning. Random Flip Augmentation and Random Zoom Augmentations were performed to achieve this.

The entire data pre-processing pipeline can be summarised as follows. for an input image $I$, the first step is to segment the Knee region, 
\begin{equation}
    I \xrightarrow{\text{Segmentation}} I_s,
\end{equation}
then apply CLAHE to adjust the contrast, sharpness, and edge definition,
\begin{equation}
    I_s \xrightarrow{\text{CLAHE}} I_c,
\end{equation}
and finally apply random augmentations to obtain the final preprocessed image $I_f$,
\begin{equation}
    I_c \xrightarrow[\text{Random Flip}]{\text{Random Zoom}} I_f.
\end{equation}

After preprocessing all the images from the dataset, the dataset was divided into 02 sets, i.e., training and validation.

\subsection{Fine-tuning CNNs}

The next step is fine-tuning each pre-trained CNN model on the pre-processed OAI dataset for both multiclass and binary classification. The following CNNs: MobileNetV2, YOLOv8, EfficientNet, DenseNet201, CvT, and ResNet50 were used. These models are pre-trained on the ImageNet1K Dataset.

% about each of the model 1-2 lines
Though MobileNetV2 (53-layer deep CNN) was specifically designed for applications in the field of mobile and embedded vision, it employs inverted residual blocks with linear bottlenecks to strike an optimal balance between model accuracy and computational efficiency.

YOLOv8 is used for real-time object detection and classification. For detection, it uses a look-once strategy, dividing the entire image into spatially separated bounding boxes and associates a class probability with each bounding box. Further, it frames the classification problem as a regression problem of each bounding box and its associated class probabilities.

EfficientNet employs a compound scaling technique to adjust network depth, width, and input resolution based on the specific task at hand. It uses a baseline network (EfficientNet-B0) that is optimized through neural architecture search to achieve superior performance and efficiency across a wide range of tasks, which include Image segmentation, Object Detection, and even language processing.

ResNet50 was designed to address the problem of vanishing gradients that are commonly observed in deep neural networks. This issue was addressed by incorporating residual connections, enabling the network to learn residual functions rather than direct mappings. This approach ensures that gradients flow more effectively through deep networks, enabling the model to optimize more efficiently and achieve better performance in complex tasks.

DenseNet201 or Densely Connected Convolutional Networks architecture connects each layer to every preceding layer within dense blocks, enabling feature reuse and improving gradient flow. This design enhances learning and mitigates issues like overfitting and vanishing gradients and results in fewer parameters.

CvT improves upon existing transformer-based image classification models by integrating convolutional layers, enhancing both performance and efficiency. CvT introduces convolutional token embedding and convolutional Transformer blocks, combining CNN feature extraction properties with Transformer attention-based advantages.

\begin{figure}
    \centering
    \includegraphics[width=0.99\linewidth]{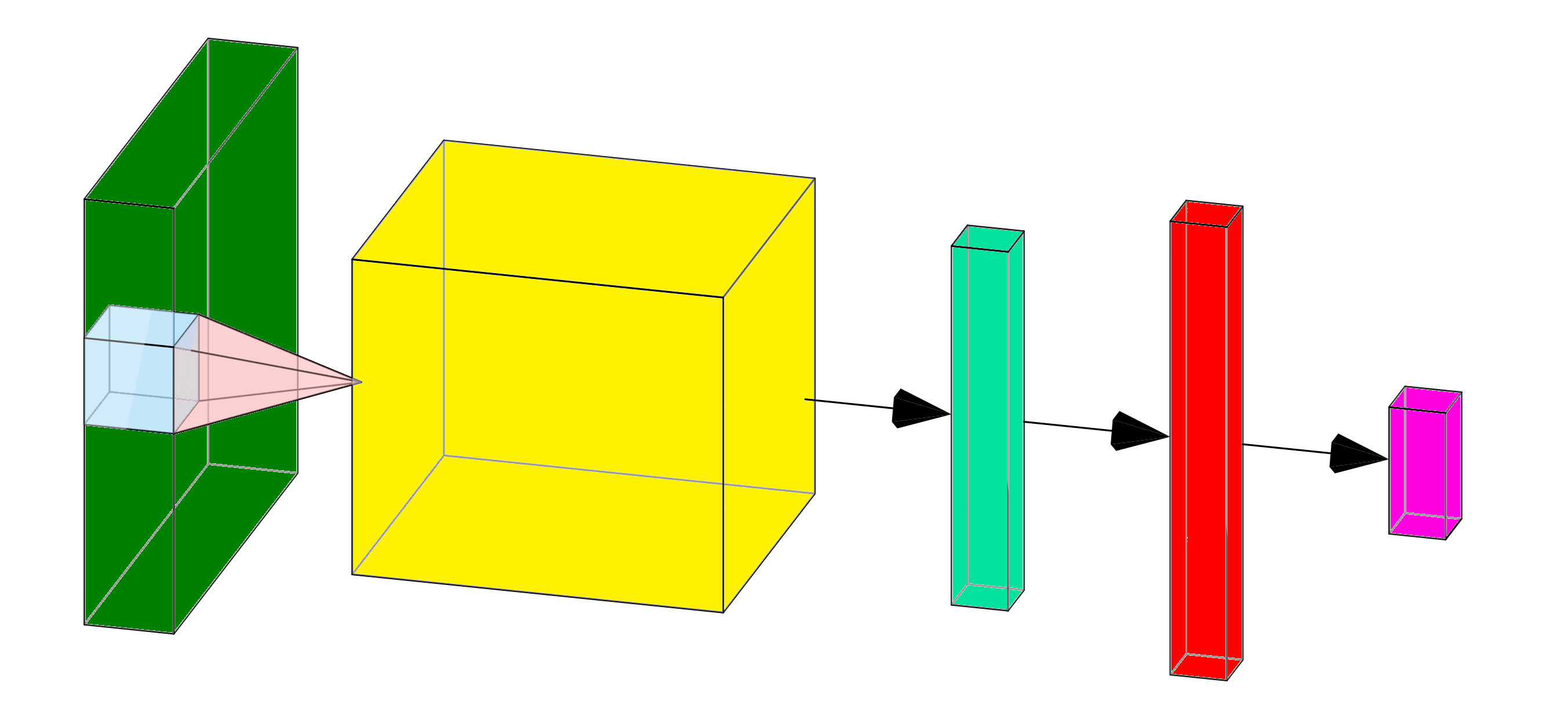}
    \caption{The final model architecture which is fine-tuned on the OAI \cite{OAI} dataset. The green layer represents the input image $I$. The yellow block represents the pre-trained CNN Backbone (\ref{eq:cnn}), the light green layer represents the vector output of 2D global average pooling (\ref{eq:globalpool}) followed by Dense Layer (red color) with ReLU activation (\ref{eq:dense1}). For multiclass classification (Grading), the last Dense Layer has 5 Nodes with Softmax activation, and for Binary Classification (Detection), the last Dense Layer has 1 Node with Sigmoid activation.}
    \label{fig:nn_2__1_}
\end{figure}

% how we modified the model architecture before fine-tuning
The final model architecture, which is fine-tuned on the pre-processed dataset, is shown in Fig. \ref{fig:nn_2__1_}. In this final model, the pre-trained models form the backbone, after which a GlobalAveragePooling2D layer is added to flatten the output and extract the features. This is followed by a Dense layer with 320 Nodes and Rectified Linear Unit (ReLU) activation, followed by a Dropout Layer for more robust training. The final layer of the model depends on the task. %For \enquote{multiclass classification}, a dense layer with 5 nodes and a Softmax activation function is added, while for \enquote{binary classification}, a dense layer with 1 node and a Sigmoid activation function is included.

During inference, for an input image $I$, the first step is to apply the fine-tuned CNN backbone,
\begin{equation}
        I \xrightarrow{\text{CNN Backbone}} I_c,
        \label{eq:cnn}
\end{equation}
where $I_c$ represents the feature map extracted by the fine-tuned CNN. The next step is to apply GlobalAveragePooling2D, which makes the CNN more robust for feature extraction,
\begin{equation}
    I_c \xrightarrow{\text{GlobalAveragePooling2D}} \textbf{X},
    \label{eq:globalpool}
\end{equation}
the intermediate vector $\textbf{X}$ is then passed onto a 320 Nodes Dense Layer with ReLU activation to get another intermediate vector $\textbf{Z}$,
\begin{equation}
        \textbf{X} \xrightarrow[\text{ReLU}]{\text{Dense Layer [320 Nodes]}} \textbf{Z} .
        \label{eq:dense1}
\end{equation}

The final transformation applied on $\textbf{Z}$ is task dependent. For multiclass classification, 
\begin{equation}
    \textbf{Z} \xrightarrow[\text{Softmax}]{\text{Dense Layer [5 Nodes]}} \textbf{Y},
\end{equation}
and for binary classification,
\begin{equation}
    \textbf{Z} \xrightarrow[\text{Sigmoid}]{\text{Dense Layer [1 Node]}} \textbf{Y},
\end{equation}
where $\textbf{Y}$ is the final output vector that is used for classification.

% \begin{wrapfigure}[]{l}{0.45\textwidth} % Adjust the height with [15] and width with {0.45\textwidth}
%     \centering
%     \includegraphics[scale=0.2]{Archi-KOA.png}
%     \caption{Fine-tuning on pre-trained models}
% \end{wrapfigure}

An optimizer, named \enquote{Stochastic Gradient Descent (SGD)}, was used to fine-tune the models with momentum \cite{momentum}. It maintains a moving average of the slope of the loss function $V_{t-1}$ weighted by the momentum coefficient $\beta$ and uses it in conjunction with the current gradient $\nabla_wL(W, X,y)$ to calculate the final update value $V_t$,
\begin{equation}
    V_t = \beta*V_{t-1} + \alpha*\nabla_wL(W,X,y),\\    
\end{equation}
where $\alpha$ is the learning rate. 

This calculated $V_t$ is then used to update the parameters of the network,
\begin{equation}
    W = W - V_t.
\end{equation}

In multiclass classification, a class-weighted Categorical Cross Entropy (CE) loss function is used \cite{categorical cross entropy}, and in binary classification, a class-weighted Binary CE loss function is used. CE Loss quantifies the difference between predicted probabilities and true categorical labels. Class weights are used to impose a greater penalty on misclassifications of the minority class. The objective is to enhance the model's sensitivity for the minority class by increasing the cost associated with incorrectly classifying instances from that class. The class-weighted categorical CE loss $L_{CE}$ for 1 sample over $N$ classes is calculated using,
\begin{equation}
    L_{CE} = - \sum_{i_c=1}^{N} \alpha_{i_c} * y_{i_c} * log(\hat{y_{i_c}}),
    \label{eq:CE}
\end{equation}
where $\alpha_{i_c}$ is the class weight associated with the $i_c^{\text{th}}$ class, $y_{i_c}$ is an indicator variable, such that
\begin{equation}
    y_{i_c} = \begin{cases}
        1 & \text{If } i_c^{\text{th}} ~ \text{sample} \in i_c \text{ Class} \\
        0 & \text{else},
    \end{cases}
\end{equation}
and $\hat{y_{i_c}}$ is the associated class probability for the $i_c^{\text{th}}$ class given by the model.

\subsection{Stacked Ensemble}

After fine-tuning the CNNs, the models that had the best accuracies on the testing sets were selected as the base learners for stacked ensemble. The meta-learners used were Random Forest, CatBoost, KNN, LightGBM, and TabNet. Grid Search Cross-Validation (GridSearchCV) and Randomised Search Cross-Validation (RandomisedSearchCV) were utilized to fine-tune the hyperparameters of the meta-learners.

The stacking ensemble uses a meta-learner ($H$) which takes the predictions of ($h^b_i\text{'}s$), i.e., base learners, on the input vector $\textbf{x}$ to generate the final class probabilities ($\hat{y}$),
\begin{equation}
\hat{y} = H([ \; h^b_1(\textbf{x}) \; | \; h^b_2(\textbf{x}) \; | \; ... \; |  \; h^b_n(\textbf{x}) \; ]^T).    
\end{equation}

To make the process of training the meta-learners and fine-tuning their hyperparameters easier and faster, the entire train, validation, and test sets were passed through the base learners for multiclass and binary classification to get their class probability prediction. After hyperparameter tuning, the best classifier was trained on the train set and used to generate the accuracy score, balanced accuracy score, and AUC values on the train, validation, and test sets.

\section{Experimental Results, Validations and Discussions}
\label{sec: Experimental Results, Validations, and Discussions}

This section highlights the various settings and values of hyperparameters used while fine-tuning the CNNs and while training the meta-learners for the final Stacking Ensemble model. Further, they contain the fine-tuning results of CNNs and the final results of the meta-learners.

\subsection{Dataset and Links}

\begin{figure}
    \centering
    \includegraphics[width=0.99\linewidth]{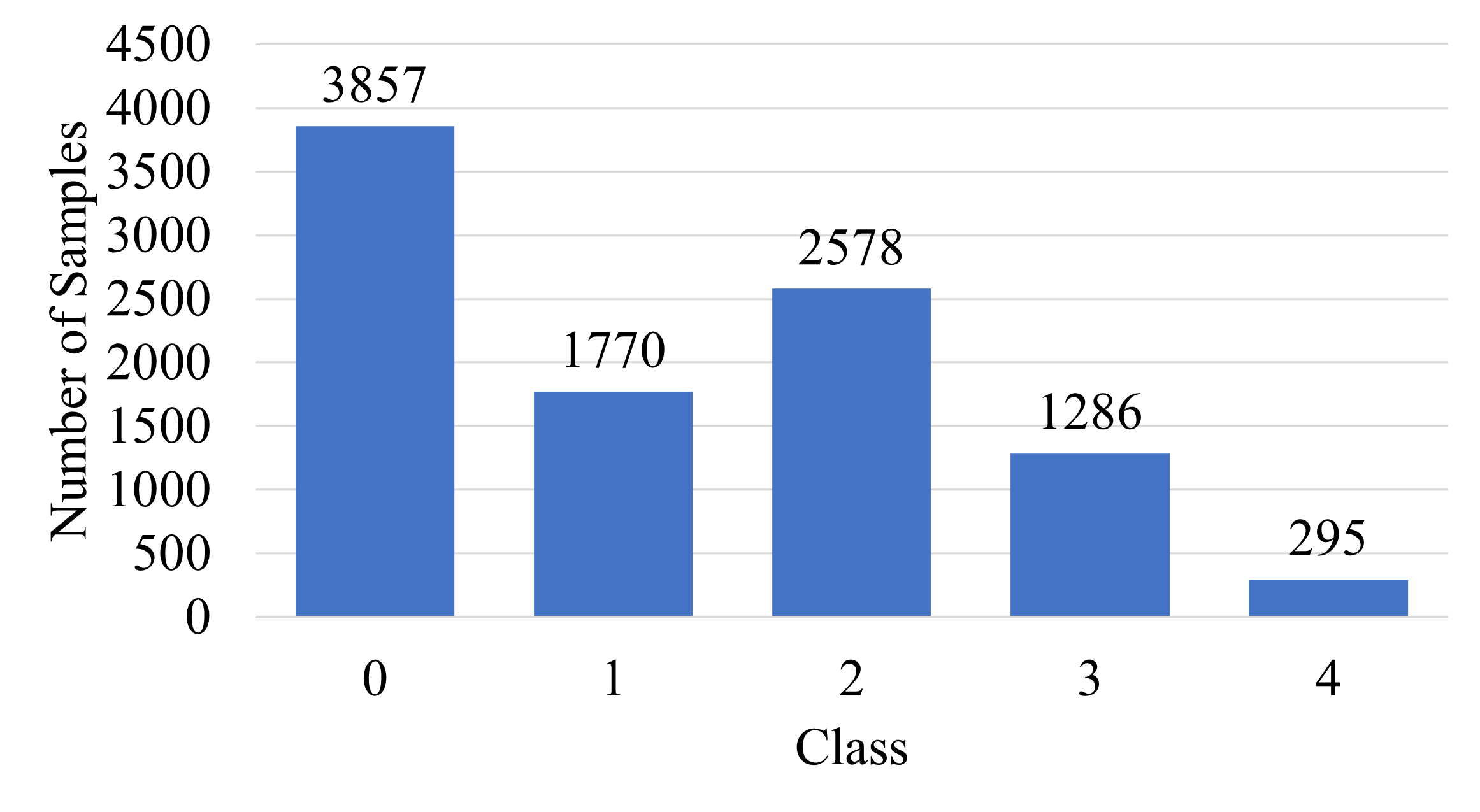}
    \caption{Class-wise distribution of the number of samples in the OAI \cite{OAI} dataset. Here, 0, 1, 2, 3, and 4 represent \enquote*{No Disease}, \enquote*{Doubtful}, \enquote*{Minimal}, \enquote*{Moderate}, and \enquote*{Severe} classes, respectively.}
    \label{fig:class-wise-distribution-of-number-of-sample}
\end{figure}

The source of the Knee X-ray images is the OAI dataset, which is most extensively used in extant literature. The Osteoarthritis Initiative, also known as OAI, is a nationwide research study sponsored by the National Institute of Health. The public source clinical Knee X-Ray data released as part of this study is used for training models. This dataset has 9786 X-ray images, which were partitioned into the train and validation sets. The dataset is highly imbalanced, with a high degree of variation in the number of samples in each class, as shown in Fig. \ref{fig:class-wise-distribution-of-number-of-sample}. This was countered by using a class-weighted loss function.

While pre-processing the dataset, after segmentation, the images needed to be converted to grayscale images as required by CLAHE. CLAHE was used with a clipping limit of 3. Following this, augmentations were performed using Random Flips and Random Zoom (0.1) to make the model more robust. Additionally, the images were resized appropriately based on the input size of the CNN.

The dataset has 5 classes of images based on the KL scale. The dataset was used as is for multiclass classification. For the task of binary classification, the data of class 0 and class 1 were combined to create Class 0 (Negative Prognosis), and classes 2, 3, and 4 were combined to create Class 1 (Positive Prognosis). 

\subsection{Fine-tuning CNNs: Settings}

The models employed were pre-trained on the ImageNet1k dataset. This dataset consists of 1000 classes with over a million image samples.

For all models, a dense layer with 320 Nodes with ReLU activation was added. Further, a dropout layer with a dropout probability of 0.2 was added, which drops 20\% of the nodes randomly during training to prevent overfitting. The last layer of the model is task-specific. A dense output layer with 5 Nodes is added with Softmax activation function for multiclass classification, and a dense output layer with 1 Node is added with Sigmoid activation for binary classification.

During training, the SGD optimizer was employed with a \enquote{learning rate = 0.001}, and a \enquote{momentum = 0.9}. The loss function applied was the class-weighted Cross-Entropy (CE) loss. Each architecture was run for a varied amount of epochs:  MobileNetV2 (40 epochs), YOLOv8 (150 epochs), EfficientNet (10 epochs), DenseNet201 (9 epochs), CvT (3 epochs), and ResNet50 (5 epochs).

\subsection{Stacked Ensemble: Settings}

To select the base learners, a threshold of 50\% testing accuracy for multiclass classification and 70\% testing accuracy for binary classification was chosen to decide the models. Thus, DenseNet201, YOLOv8, and MobileNetV2 were used as base learners in the ensemble.

In multiclass classification, CatBoost had the highest testing accuracy; the hyperparameters obtained for this meta-learner were depth = 10, iterations = 100, and learning rate = 0.1. However, the KNN classifier with k = 6 surpassed CatBoost on balanced test accuracy.

In binary classification, CatBoost and KNN achieved the highest test accuracy. The final hyperparameters obtained for CatBoost were depth = 15, iterations = 100, and learning rate = 0.00005, while those for KNN were k = 4.

\subsection{Fine-tuning CNNs: Results}

\begin{figure}
    \centering
    \includegraphics[width=0.95\linewidth]{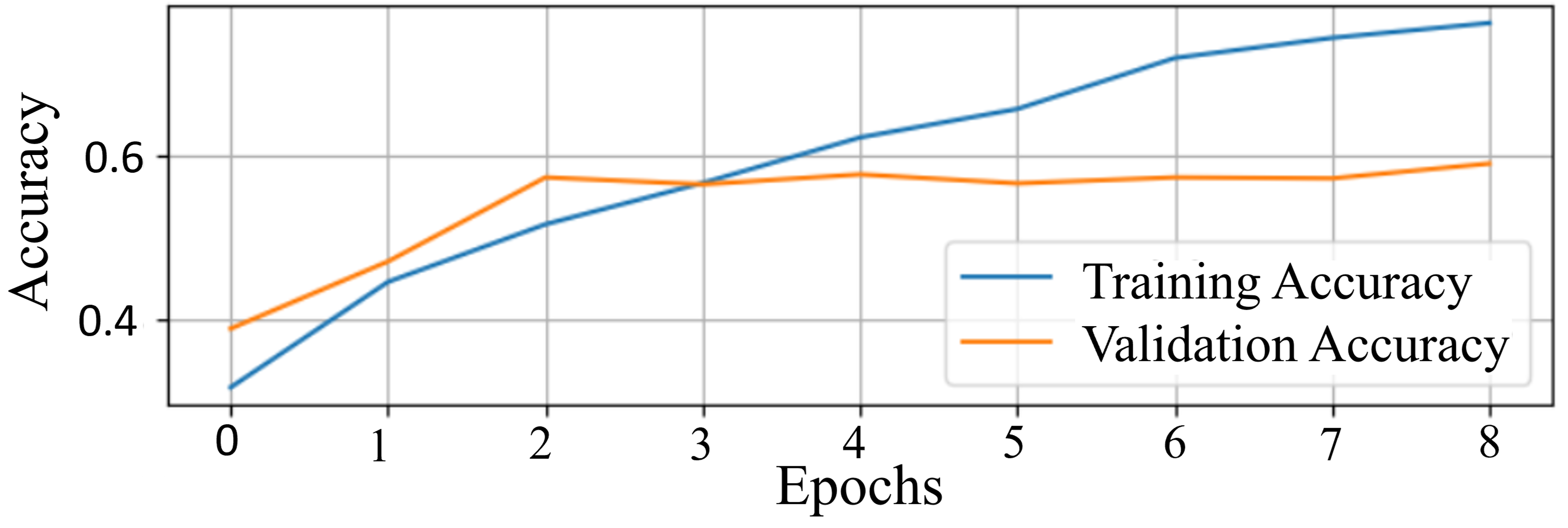}
    \caption{Observed trends for DenseNet201 \cite{densenet} fine-tuning for Multiclass Classification.}
    \label{fig:DensenetMulticlassTraining}
\end{figure}

\begin{figure}
    \centering
    \includegraphics[width=0.95\linewidth]{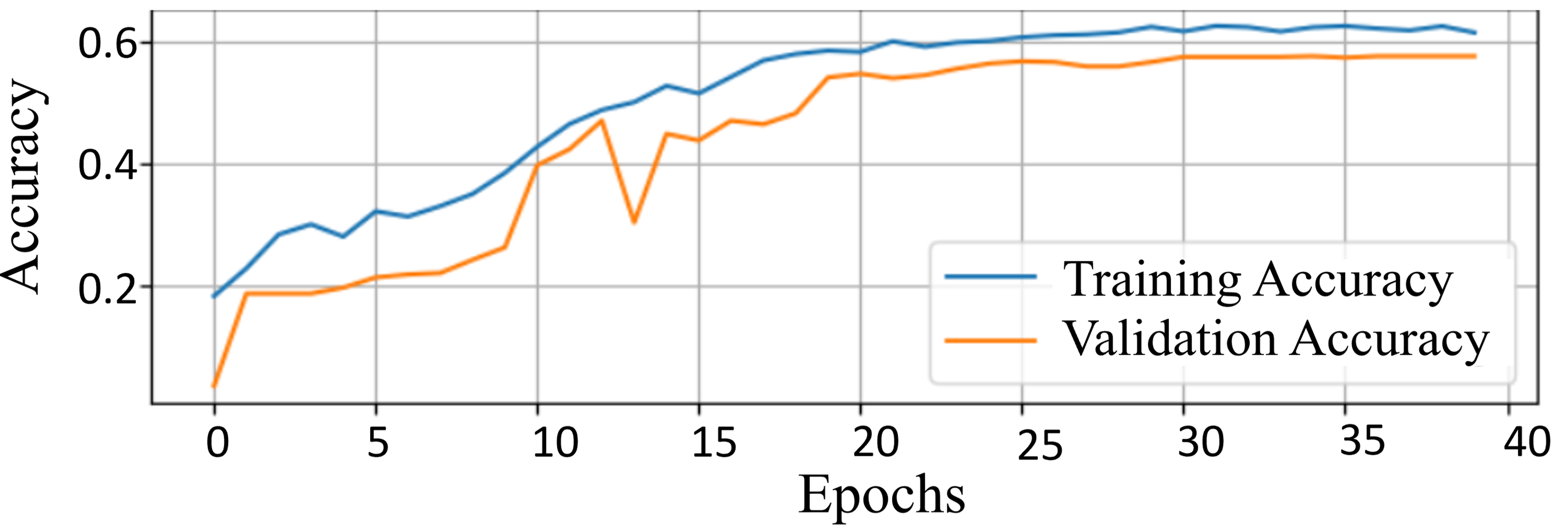}
    \caption{Observed trends for MobileNetV2 \cite{MobileNetv2} fine-tuning for Multiclass Classification.}
    \label{fig:MobilnetMulticlassFinetuning}
\end{figure}

For the task of multiclass classification, DenseNet201, YOLOv8, and MobileNetV2 performed well with a testing accuracy above 0.5 (out of 5 classes). EfficientNet, CvT, and ResNet50 did not give satisfactory results, with testing accuracies of 0.327, 0.03, and 0.258, respectively. The accuracy and AUC on the different sets (that includes \enquote{train}, \enquote{validation}, and \enquote{test}) are given in Table \ref{table:MULclassification_performance}. The observation trends against the epochs for fine-tuning of \enquote{DenseNet201} and \enquote{MobileNetV2} are given in Figs. \ref{fig:DensenetMulticlassTraining} and \ref{fig:MobilnetMulticlassFinetuning}, respectively.

\begin{table}
    \centering
    \caption{Fine-tuning results of Multiclass Classification.}
    \label{table:MULclassification_performance}
    \begin{tabular}{|c|c|c|c|c|}
        \hline
        \textbf{Model} & \textbf{Metric} & \textbf{Train} & \textbf{Val} & \textbf{Test} \\
        \hline
        \hline
        \multirow{2}{*}{DenseNet201  \cite{densenet}} & Accuracy & \textbf{0.823} & \textbf{0.582} & \textbf{0.673} \\
        \cline{2-5}
         & AUC & \textbf{0.967} & \textbf{0.849} & \textbf{0.898} \\
        \hline
        \multirow{2}{*}{YOLOv8  \cite{YOLO}} & Accuracy & 0.687 & 0.567 & 0.631 \\
        \cline{2-5}
         & AUC & 0.908 & 0.823 & 0.876 \\
        \hline
        \multirow{2}{*}{MobileNetV2  \cite{MobileNetv2}} & Accuracy & 0.613 & 0.556 & 0.575 \\
        \cline{2-5}
         & AUC & 0.84 & 0.8 & 0.817 \\
        \hline
    \end{tabular}
    \caption*{\small Values in bold signify the highest magnitude for that metric.}
\end{table}

\begin{table}
    \centering
    \caption{Fine-tuning results of Binary Classification.}
    \label{table:BINclassification_performance}
    \begin{tabular}{|c|c|c|c|c|}
        \hline
        \textbf{Model} & \textbf{Metric} & \textbf{Train} & \textbf{Val} & \textbf{Test} \\
        \hline
        \hline
        \multirow{2}{*}{DenseNet201  \cite{densenet}} & Accuracy & 0.847 & 0.798 & 0.796 \\
        \cline{2-5}
         & AUC & 0.936 & 0.86 & 0.881 \\
        \hline
        \multirow{2}{*}{YOLOv8  \cite{YOLO}} & Accuracy & 0.813 & 0.781 & 0.793 \\
        \cline{2-5}
         & AUC & 0.899 & 0.841 & 0.876 \\
        \hline
        \multirow{2}{*}{MobileNetV2  \cite{MobileNetv2}} & Accuracy & \textbf{0.94} & \textbf{0.821} & \textbf{0.856} \\
        \cline{2-5}
         & AUC & \textbf{0.982} & \textbf{0.892} & \textbf{0.945} \\
        \hline
    \end{tabular}
    \caption*{\small Values in bold signify the highest magnitude for that metric.}
\end{table}

\begin{figure}
    \centering
    \includegraphics[width=0.95\linewidth]{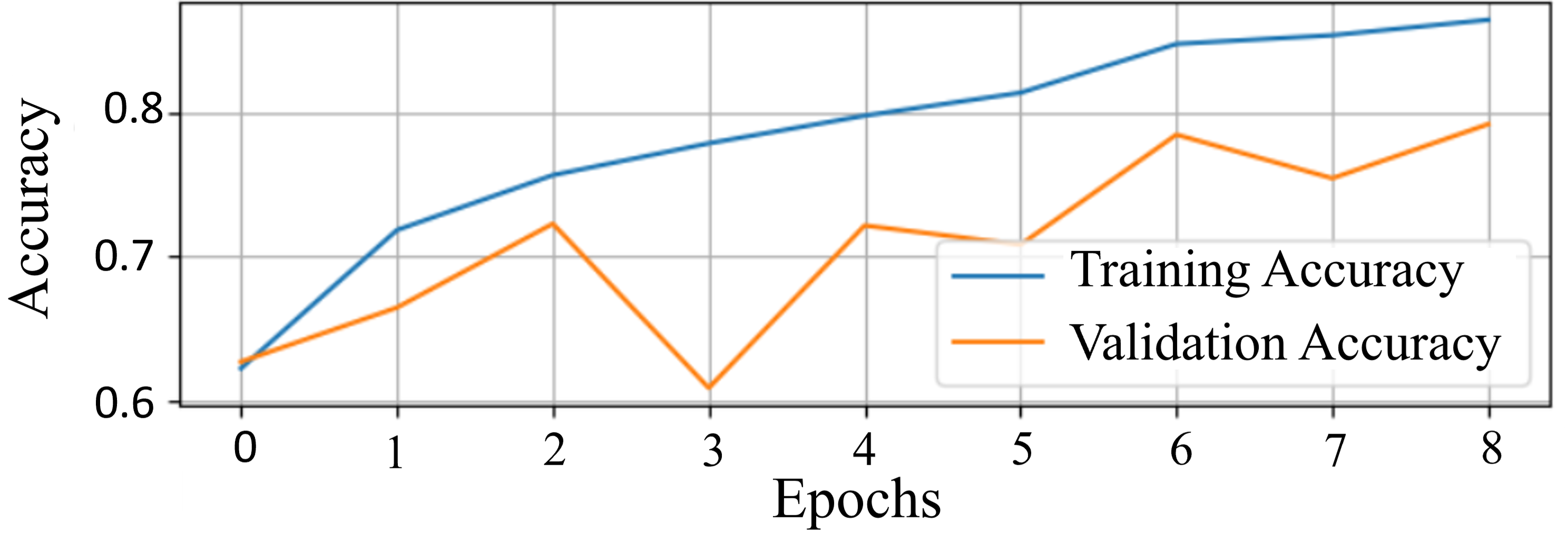}
    \caption{Observed trends for DenseNet201 \cite{densenet} fine-tuning for Binary Classification.}
    \label{fig:DensenetBinaryFinetuning}
\end{figure}

\begin{figure}
    \centering
    \includegraphics[width=0.95\linewidth]{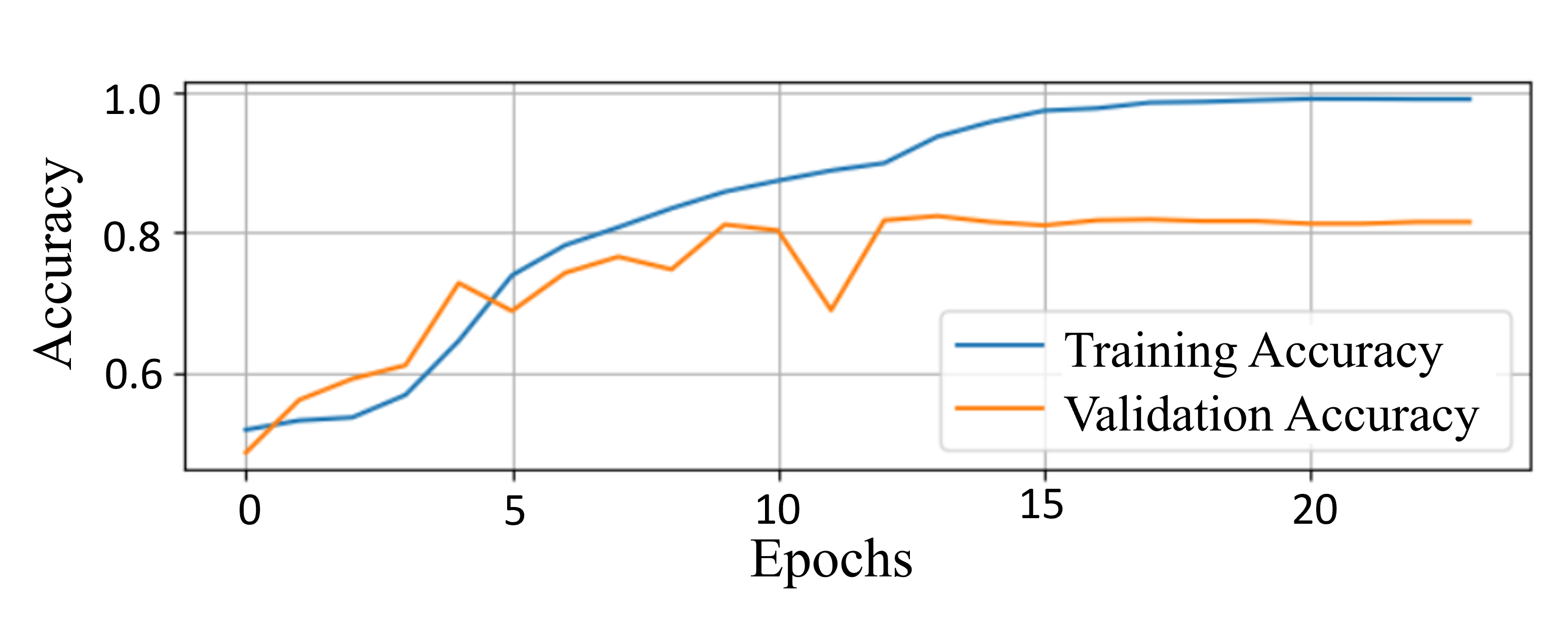}
    \caption{Observed trends for MobileNetV2 \cite{MobileNetv2} fine-tuning for Binary Classification.}
    \label{fig:MobilenetBinaryFinetuning}
\end{figure}

For the task of binary classification, again, MobileNetV2, YOLOv8, and DenseNet201 performed the best, with testing accuracies above 0.75. The other models (EfficientNet, CvT, and ResNet50) did not give satisfactory results, with testing accuracies of 0.582, 0.517, and 0.464, respectively. The accuracy and AUC on the different sets (that includes \enquote{train}, \enquote{validation}, and \enquote{test}) are given in Table \ref{table:BINclassification_performance}. The observation trends against the epochs for fine-tuning of \enquote{DenseNet201} and \enquote{MobileNetV2} are given in Figs. \ref{fig:DensenetBinaryFinetuning} and \ref{fig:MobilenetBinaryFinetuning}, respectively.

\subsection{Stacked Ensemble: Results}

In multiclass classification, CatBoost yielded the highest testing accuracy of 71.1\% and the highest testing AUC of 0.912. CatBoost achieved a balanced accuracy of 72\% on the test set, while the KNN classifier (k=6) achieved a balanced accuracy of 73\% on the test set. All the meta-learners performed reasonably well, attaining test accuracies greater than 70\% and test AUC greater than 0.87. The accuracy, balanced accuracy, and AUC on the train, validation, and test datasets are given in Table \ref{table:meta-learners_performance}.

\begin{table}
    \centering
    \caption{Performance of various meta-learners in multiclass classification.}
    \label{table:meta-learners_performance}
    \resizebox{\columnwidth}{!}{
    \begin{tabular}{|c|c|c|c|c|}
        \hline
        \textbf{Meta-learner} & \textbf{Metric} & \textbf{Train} & \textbf{Val} & \textbf{Test} \\
        \hline
        \hline
        \multirow{3}{*}{Random Forest \cite{randomforest}} & Accuracy & \textbf{0.941} & \textbf{0.738} & 0.707 \\
        \cline{2-5}
        & Balanced Accuracy & \textbf{0.948} & \textbf{0.731} & 0.724 \\
        \cline{2-5}
        & AUC & \textbf{0.997} & \textbf{0.964} & 0.909 \\
        \hline
        \multirow{3}{*}{CatBoost \cite{catboost}} & Accuracy & 0.912 & 0.701 & \textbf{0.711} \\
        \cline{2-5}
        & Balanced Accuracy & 0.921 & 0.699 & 0.720 \\
        \cline{2-5}
        & AUC & 0.988 & 0.917 & \textbf{0.912} \\
        \hline
        \multirow{3}{*}{KNN \cite{knn}} & Accuracy & 0.879 & 0.657 & 0.709 \\
        \cline{2-5}
        & Balanced Accuracy & 0.890 & 0.658 & \textbf{0.730} \\
        \cline{2-5}
        & AUC & 0.982 & 0.919 & 0.875 \\
        \hline
        \multirow{3}{*}{LightGBM \cite{lgbm}} & Accuracy & 0.863 & 0.613 & 0.705 \\
        \cline{2-5}
        & Balanced Accuracy & 0.879 & 0.629 & 0.712 \\
        \cline{2-5}
        & AUC & 0.973 & 0.860 & 0.906 \\
        \hline
        \multirow{3}{*}{TabNet \cite{tabnet}} & Accuracy & 0.873 & 0.607 & 0.708 \\
        \cline{2-5}
        & Balanced Accuracy & 0.891 & 0.626 & 0.723 \\
        \cline{2-5}
        & AUC & 0.978 & 0.811 & 0.889 \\
        \hline
    \end{tabular}
    }
    \caption*{\small Values in bold signify the highest magnitude for that metric.}
\end{table}

\begin{table}
    \centering
    \caption{Performance of various meta-learners in binary classification.}
    \label{table:meta-learners_performance_updated}
    \resizebox{\columnwidth}{!}{
    \begin{tabular}{|c|c|c|c|c|}
        \hline
        \textbf{Meta-learner} & \textbf{Metric} & \textbf{Train} & \textbf{Val} & \textbf{Test} \\
        \hline
        \hline
        \multirow{3}{*}{Random Forest \cite{randomforest}} & Accuracy & \textbf{0.991} & \textbf{0.958} & 0.877 \\
        \cline{2-5}
        & Balanced Accuracy & \textbf{0.990} & \textbf{0.956} & 0.872 \\
        \cline{2-5}
        & AUC & \textbf{1.000} & \textbf{0.996} & 0.939 \\
        \hline
        \multirow{3}{*}{CatBoost \cite{catboost}} & Accuracy & 0.959 & 0.828 & \textbf{0.879} \\
        \cline{2-5}
        & Balanced Accuracy & 0.955 & 0.819 & \textbf{0.875} \\
        \cline{2-5}
        & AUC & 0.986 & 0.904 & \textbf{0.945} \\
        \hline
        \multirow{3}{*}{KNN \cite{knn}} & Accuracy & 0.959 & 0.855 & 0.879 \\
        \cline{2-5}
        & Balanced Accuracy & 0.954 & 0.843 & \textbf{0.875} \\
        \cline{2-5}
        & AUC & 0.995 & 0.953 & 0.913 \\
        \hline
        \multirow{3}{*}{LightGBM \cite{lgbm}} & Accuracy & 0.960 & 0.828 & 0.874 \\
        \cline{2-5}
        & Balanced Accuracy & 0.957 & 0.820 & 0.870 \\
        \cline{2-5}
        & AUC & 0.987 & 0.906 & 0.945 \\
        \hline
        \multirow{3}{*}{TabNet \cite{tabnet}} & Accuracy & 0.865 & 0.757 & 0.806 \\
        \cline{2-5}
        & Balanced Accuracy & 0.880 & 0.778 & 0.823 \\
        \cline{2-5}
        & AUC & 0.984 & 0.902 & 0.945 \\
        \hline
    \end{tabular}
    }
    \caption*{\small Values in bold signify the highest magnitude for that metric.}
\end{table}

In binary classification, both CatBoost achieved the highest test accuracy of 87.9\%, showcasing the strongest performance. CatBoost and KNN also achieved the highest balanced accuracy on the test set of 87.5\%. TabNet got the lowest test accuracy of 80.6\%. Notably, all the meta-learners performed exceptionally well by attaining test accuracies greater than 80\% and test AUCs greater than 0.91. The accuracy, balanced accuracy, and AUC on the train, validation, and test datasets are given in Table \ref{table:meta-learners_performance_updated}.

\section{Conclusions and Future Scopes}
\label{sec:Conclusion and Future Scopes}

This paper presents a stacked ensemble model for severity grading of Knee Osteoarthritis (KOA) using the OAI dataset. The model combines fine-tuned CNNs to create a more effective image classification approach compared to individual CNNs. As demonstrated in Section \ref{sec: Experimental Results, Validations, and Discussions}, the ensemble models successfully classified the majority of the 9786 images from the OAI dataset, which were distributed across five imbalanced classes, achieving superior accuracy compared to previous studies in the literature. The highest balanced testing accuracies attained were 73\% for KOA grading and 87.5\% for KOA detection.

In the future, other ensemble techniques, such as bagging, can be used to reduce variance and further improve test accuracy. The meta-learners can also be trained on the feature extracted by the fine-tuned models instead of their classification outcomes. Moreover, transformer-based CNNs can be incorporated by appropriately fine-tuning with better computational resources and a larger set of meta-learners can be explored as well.

\end{document}